\documentclass{article}

\usepackage{PRIMEarxiv}

\usepackage[utf8]{inputenc} 
\usepackage[T1]{fontenc}    
\usepackage{hyperref}       
\usepackage{url}            
\usepackage{booktabs}       
\usepackage{amsfonts}       
\usepackage{nicefrac}       
\usepackage{microtype}      
\usepackage{lipsum}
\usepackage{fancyhdr}       
\usepackage{graphicx}       
\graphicspath{{media/}}     

\title{Deep learning universal crater detection using Segment Anything Model (SAM)
}

\author{
  Iraklis Giannakis$^{1}$, Anshuman Bhardwaj$^{1}$, Lydia Sam$^{1}$ and Georgios Leontidis$^{2}$ \\
  School of Geosciences$^{1}$, Interdisciplinary Centre for Data and AI $^{2}$ \\
  University of Aberdeen \\
  Aberdeen, United Kingdom\\
  \texttt{\{iraklis.giannakis, georgios.leontidis, anshuman.bhardwaj,  lydia.sam \}@abdn.ac.uk} \\
 
}

\begin{document}
\maketitle

\begin{abstract}

Craters are amongst the most important morphological features in planetary exploration. To that extent, detecting, mapping and counting craters is a mainstream process in planetary science, done primarily manually, which is a very laborious and time-consuming process. Recently, machine learning (ML) and computer vision have been successfully applied for both detecting craters and estimating their size. Existing ML approaches for automated crater detection have been trained in specific types of data e.g. digital elevation model (DEM), images and associated metadata for orbiters such as the Lunar Reconnaissance Orbiter Camera (LROC) etc.. Due to that, each of the resulting ML schemes is applicable and reliable only to the type of data used during the training process. Data from different sources, angles and setups can compromise the reliability of these ML schemes. In this paper we present a universal crater detection scheme that is based on the recently proposed Segment Anything Model (SAM) from META AI. SAM is a prompt-able segmentation system with zero-shot generalization to unfamiliar objects and images without the need for additional training. Using SAM we can successfully identify crater-looking objects in any type of data (e,g, raw satellite images Level-1 and 2 products, DEMs etc.) for different setups (e.g. Lunar, Mars) and different capturing angles. Moreover, using shape indexes, we only keep the segmentation masks of crater-like features. These masks are subsequently fitted with an ellipse, recovering both the location and the size/geometry of the detected craters.

\end{abstract}

\keywords{Segmentation\and Segment Anything Model \and SAM \and Crater Detection \and Machine Learning \and AI \and Planetary Science \and Terrestrial Planets \and Computer Vision}

\section{Introduction}
Impact craters are circular-elliptical depressions in planetary surfaces caused by the impact of meteorites, asteroids or comets \cite{Melosh:1989}. The size and the shape of craters depend on numerous factors, giving rise to a plethora of crater types with varying diameters ranging from a few meters to hundreds of kilometers \cite{Salamunicar:2012}. Impact craters are amongst the most important morphological features of planetary exploration \cite{McSween:2019}, and they have been extensively used for inferring the composition and structure of celestial bodies \cite{Lemelin:2019}. Craters act as natural excavation sites to study stratigraphy, strata and stratification, providing pivotal information for the geology and landscape evolution of the planet \cite{Huang:2018}. The distribution of crater sizes has also been widely applied for estimating the age of planetary surfaces, through calculation of crater size-frequency distribution (CSFD) and chronostratigraphy \cite{Hartman:2001}. Apart from CSFD, the shape and the erosion of the crater have also been shown to have a causal relationship with the age of the impact \cite{Yang:2020}. Moreover, impact craters can potentially become important sources of natural resources, such as frozen water in the permanently shadowed craters on the Moon \cite{Glaser:2014}. Lastly, crater spatial distributions are important for terrain-relative navigation \cite{Emami:2019, Downes:2020, Silvestrini:2022}; and for selecting landing sites for spacecrafts and landers \cite{Grant:2018}. 

To that extent, detecting and counting craters is of great importance in planetary science \cite{Silburt:2018}. Manually mapping craters via visual inspection is a very laborious and time-consuming process, that cannot be scaled up for large areas of investigation \cite{Silburt:2018}; and cannot be utilised for real-time crater detection \cite{Downes:2020}. Manual crater detection can also be subjected to human errors and biases that can lead to up to 40\% of disagreements \cite{Robbins:2014}. The above led to the development of semi-automatic crater detection algorithms (CDA) to allow for large scale real-time crater detection, and reduce human biases \cite{Silburt:2018}.

Automatic crater detection is a challenging scientific endeavour due to the wide variety of impact craters, the diversity of input data, and the level of background noise \cite{Emami:2019}. Various methodologies have been suggested for CDA over the years. From convolutional neural networks (CNN) combined with Canny edge detection \cite{Emami:2015}, to hybrid supervised-unsupervised machine learning  \cite{Emami:2019}, and using Adaboost with support vector machines for detecting craters on Mars \cite{Wetzler:2005}. In \cite{Salamuniccar:2011} 77 CDA methodologies are outlined, divided in image-based and digital elevation model (DEM)-based approaches \cite{Di:2014}. With the recent advancement of deep learning, U-nets have been successfully applied for detecting and estimating the size of impact craters. In \cite{Klear:2018}, a set of U-nets are trained using labelled photos from the Mars express mission. In \cite{Lee:2019}, topography data from Mars are used to train U-nets. U-nets have also been trained using Lunar DEM \cite{Silburt:2018} and photos from the Lunar Reconnaissance Orbiter Camera (LROC) \cite{Downes:2020}. In \cite{Yang:2020} data from Chang'E-1 and E-2 are used for training ML for detecting craters, and approximate their age based on their morphology. All these methods perform sufficiently well when applied to data similar to the ones that they have been trained for. As stated in \cite{Klear:2018}, regarding using U-nets trained for identifying craters on Mars, differences between Mars and other celestial bodies are enough to make this model unreliable outside of Mars \cite{Klear:2018}. To that extent, an algorithm that can identify craters universally, without being limited by the celestial body, data type, or measurement setup, would be highly valuable \cite{Silburt:2018}.

In the current paper we present a universal approach for identifying craters using the Segment Anything Model (SAM) \cite{Kirillov:2023}. SAM is a foundation model developed by META for computer vision and image segmentation. It was trained with over 1 billion masks to segment images in a prompt-able way allowing transfer zero-shot to new image distributions. Via this approach, regardless of the type of the data (e.g. photos, DEM, spectra, gravity etc.) or the celestial body (e.g. Moon, Mars etc.) and the measurement setup, the data will be segmented into different categories and classes. Subsequently each mask is further classified into crater and no-crater based on geometric indexes that evaluate how circular or elliptical is the investigated mask. Via numerous examples, we illustrate the effectiveness of this processing pipeline to different sets of data from different planetary bodies and measurement setups. The results highlight the potential of foundation segmentation models for crater identification and pattern recognition in planetary science in general.

\begin{figure}[t]
\includegraphics[width=8cm]{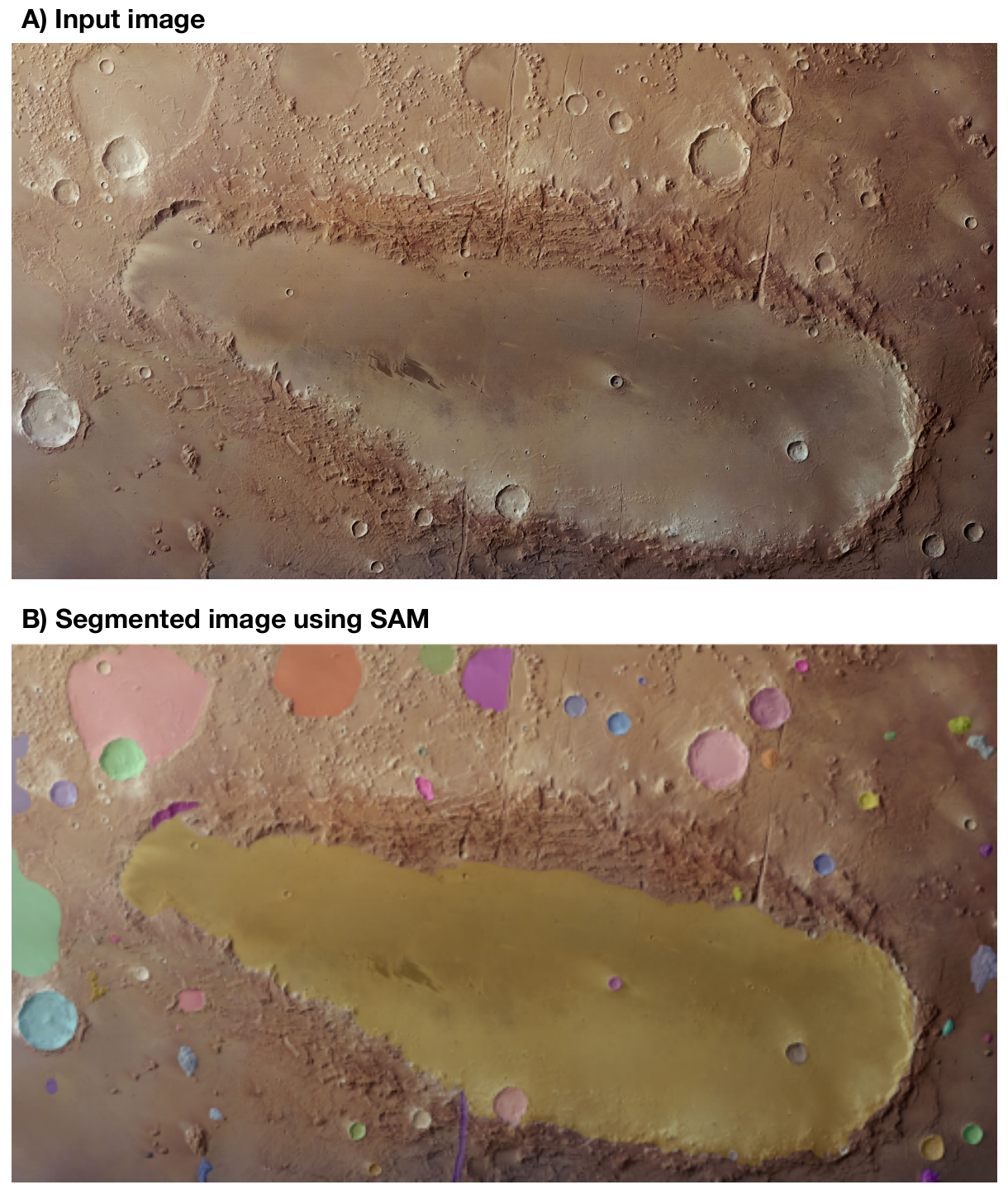}
\centering
\caption{A) Mars Express HRSC natural colour image of \textit{Ocrus Patera} B) Segmentation of the input image using Segment Anything Model (SAM) \cite{Kirillov:2023}.}
\label{F1}
\end{figure}

\begin{figure}[t]
\includegraphics[width=8cm]{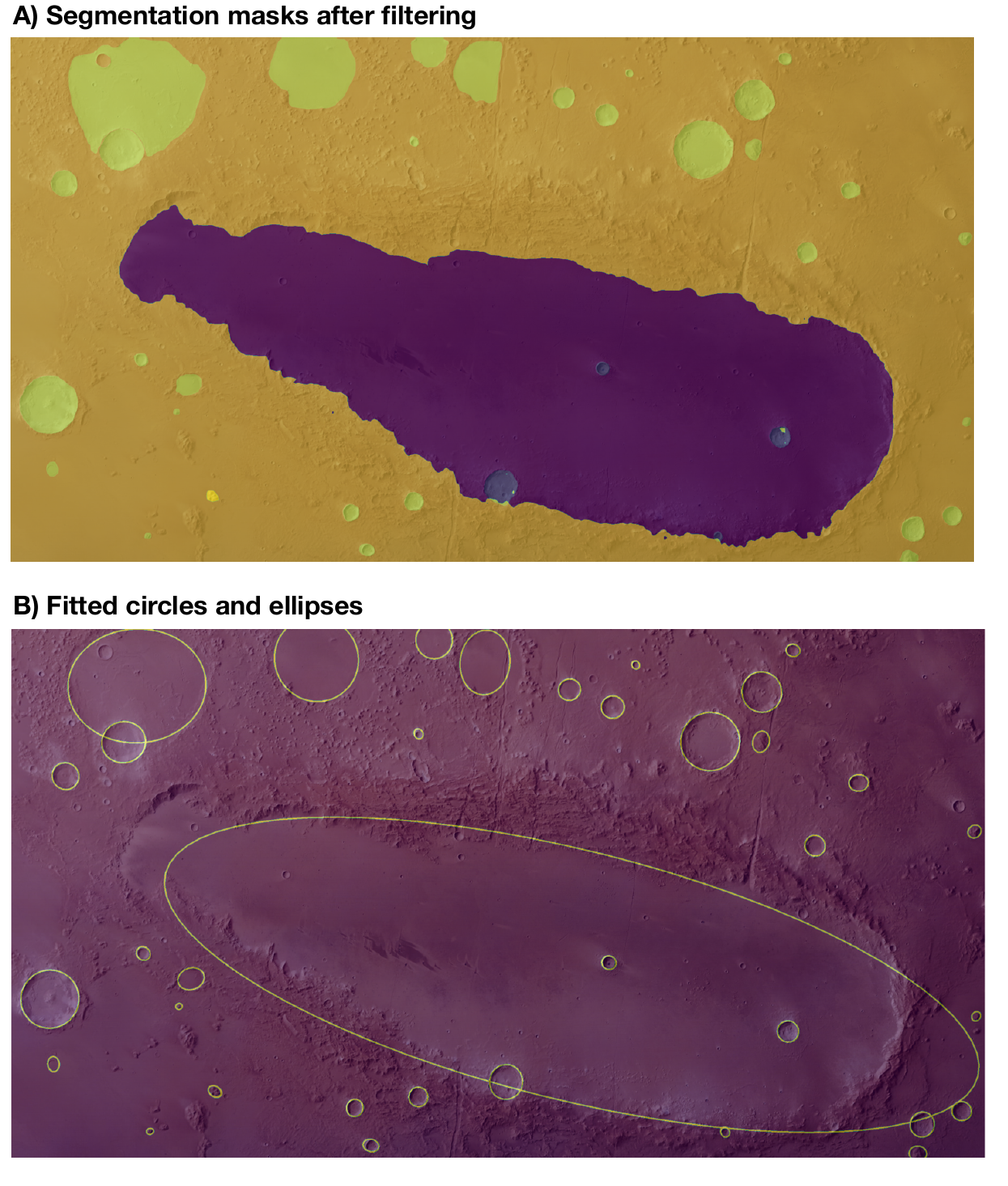}
\centering
\caption{A) The remaining segmentation masks from Fig. \ref{F1} after filtering out the non-circular/elliptical classes using geometrical indexes. B) Canny filter is applied to each one of the remaining masks, and the edges are fitted with circles and ellipses.}
\label{F2}
\end{figure}

\section{Methodology}
The processing pipeline is composed of three sequential steps. Initially, the input image undergoes segmentation using SAM \cite{Kirillov:2023}. There are no restrictions regarding the celestial body,  data type, resolution etc., any type of imagery data can be used as input. Next, each segmentation mask is analysed to determine its shape. Any masks that are not identified as circles or ellipses are filtered out, and the remaining masks are subjected to further processing to extract their boundaries and fit an ellipse to their edges. Finally, a post-processing filter is employed to eliminate any potential duplicates, artefacts, or false positives.

\subsection{Segment Anything Model (SAM)}
Image segmentation is a branch of computer vision and digital image processing aiming at clustering a given image into several segments/masks \cite{Szeliski:2011}. Numerous algorithms have been suggested for image segmentation throughout the years from using unsupervised clustering methods such as K-means \cite{Dhanachandra:2015} to histogram-based methods \cite{Qin:2011} and data coding and compression \cite{Yi:2007}. In recent years, deep learning has been extensively used for image segmentation with impressive results compared to previous approaches \cite{Farabet:2013, Chen:2018, Kim:2021, Kexin:2020, Yang:2018, Noh:2015}; and has become the standard in remote sensing segmentation in geosciences \cite{Buscombe:2022, Chen:2020, Collins:2020, Gupta:2021, Zhang:2018}, and for real time identification of objects in Martian terrain for safe rover navigation \cite{Liu:2023, Goh:2022,Liu:2023_b}.

Recently, foundation deep learning schemes \cite{Sofiiuk:2022, Qin:2022} have been developed for interactive image segmentation trained with large and diverse image databases (COCO \cite{Lin:2014}, LVIS  \cite{Gupta:2019} etc.). In April 2023, META released their own model named "Segment Anything Model" (SAM) \cite{Kirillov:2023}, a deep learning image segmentation that outperforms previous approaches. SAM has been trained in a high-quality dataset (SA-1B \cite{Kirillov:2023}) consisting of millions of images and billions of masks, significantly larger than previous databases \cite{Kirillov:2023}. SAM consists of a heavyweight image encoder that is based on Masked Autoencoders \cite{He:2022} and a pretrained Vision Transformer model \cite{Dosovitskiy:2021}. The image encoder outputs an image embedding that is subsequently used to produce the image masks. The image embedding is further enriched with a variety of input prompts such as clicks, selected boxes and text \cite{Kirillov:2023}. SAM demonstrate excellent performance in a wide range of images from databases significantly different compared to SA-1B \cite{Kirillov:2023}. The generalisation capabilities of SAM make it a potential candidate for CDA, overcoming the limitations of data-specific CDA without the need for additional training and well-labelled data.      

\begin{figure}[t]
\includegraphics[width=15cm]{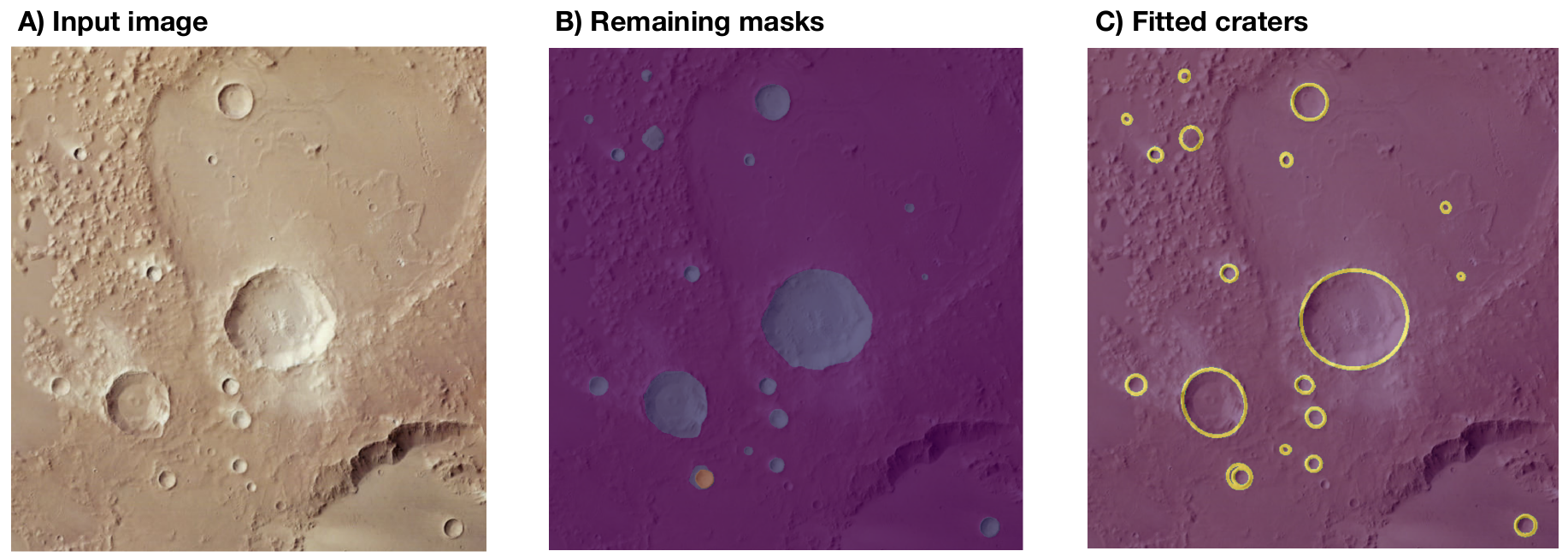}
\centering
\caption{Applying the proposed CDA to the top left area of the Mars Express HRSC natural colour image of \textit{Orcus Patera} shown in Figure \ref{F1}. A) column is the input image, B) are the remaining masks, and C) are the final fitted craters. The small undetected craters shown in Fig. \ref{F2} are now correctly identified and mapped after focusing (zoom in) the investigated area.}
\label{F3}
\end{figure}

SAM can be modified and tuned by changing its hyper-parameters, but it is generally advised to use the default ones \cite{Kirillov:2023}. The resulting outputs are outlined below:
\begin{itemize}
    \item Segmentation masks
    \item The areas of the masks in pixels
    \item The boundary box for each mask
    \item The quality of the mask (from 0 to 1)
    \item The input point that generated each mask
    \item The stability score for each mask, which is an additional quality index (from 0 to 1)
    \item The crop of the image used to generate each mask
\end{itemize}
From the above it is evident that SAM in principle can both detect and estimate the size of impact craters, since it provides direct information regarding the area of the mask and its bounding box.

\begin{figure}[t]
\includegraphics[width=15cm]{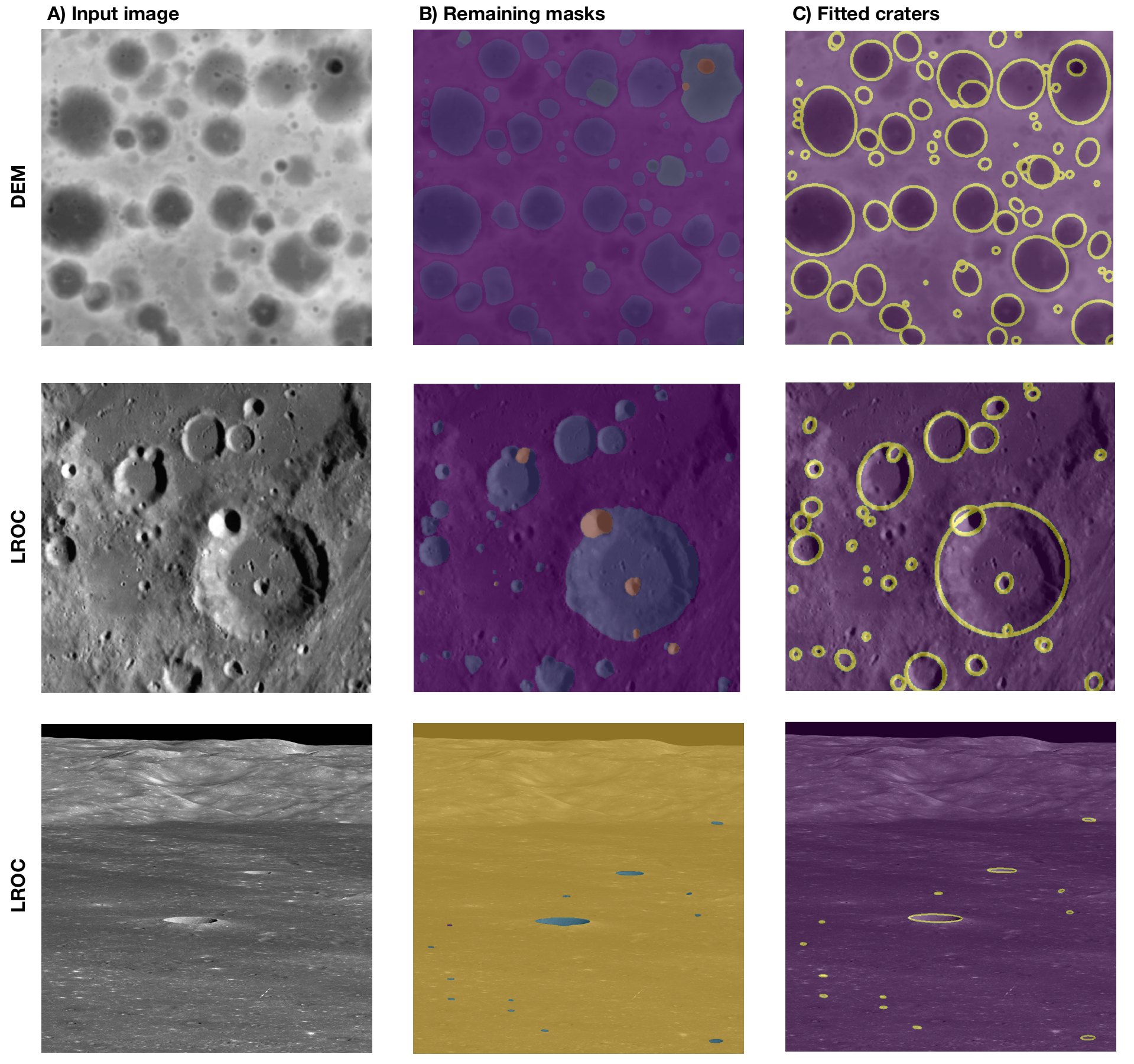}
\centering
\caption{The proposed CDA using different types of Lunar data. A) column is the input image, B) are the remaining masks after using geometrical indexes, and C) are the final fitted craters. The proposed CDA works reasonably well regardless of data type and  measurement configuration.}
\label{F4}
\end{figure}

Figure \ref{F1} illustrates an example of using SAM to a natural colour image of \textit{Ocrus Patera} taken from Mars Express High Resolution Stereo Camera (HRSC). Despite not being trained specifically for Mars Express HRSC images, SAM appears to capture all the major features in the image, which is indicative of its effectiveness. Additionally, access to dominant feature masks enables further classification based on shape and size. The next processing step involves filtering out non-circular/elliptical masks and fitting ellipses to the remaining segments.

\begin{figure}[t]
\includegraphics[width=15cm]{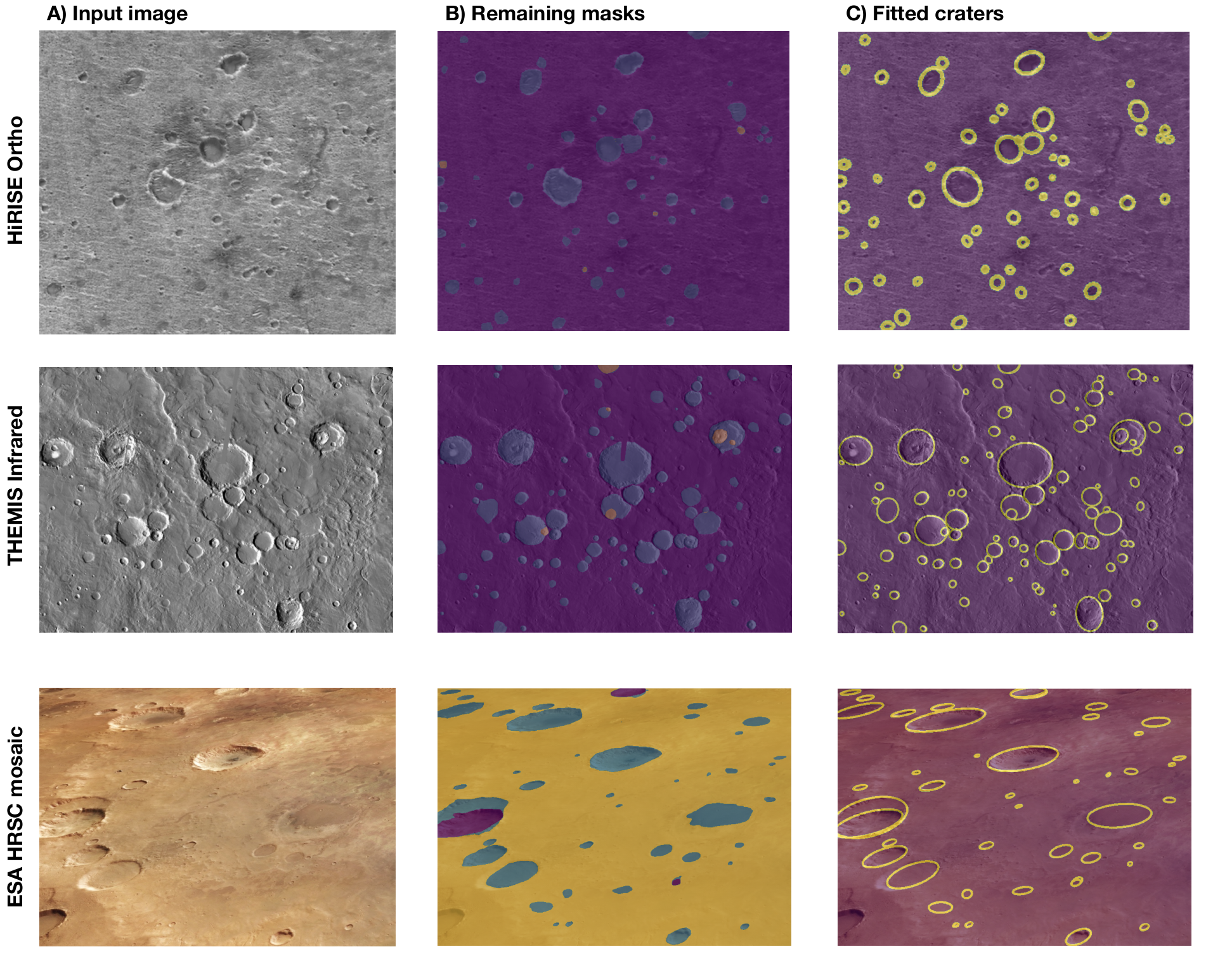}
\centering
\caption{The proposed CDA using different types of data from Mars. A) column is the input image, B) are the remaining masks after using geometrical indexes, and C) are the final fitted craters. Similar to Fig. \ref{F4}, the proposed CDA works reasonably well regardless of data type and the measurement configuration. }
\label{F5}
\end{figure}

\begin{figure}[t]
\includegraphics[width=15cm]{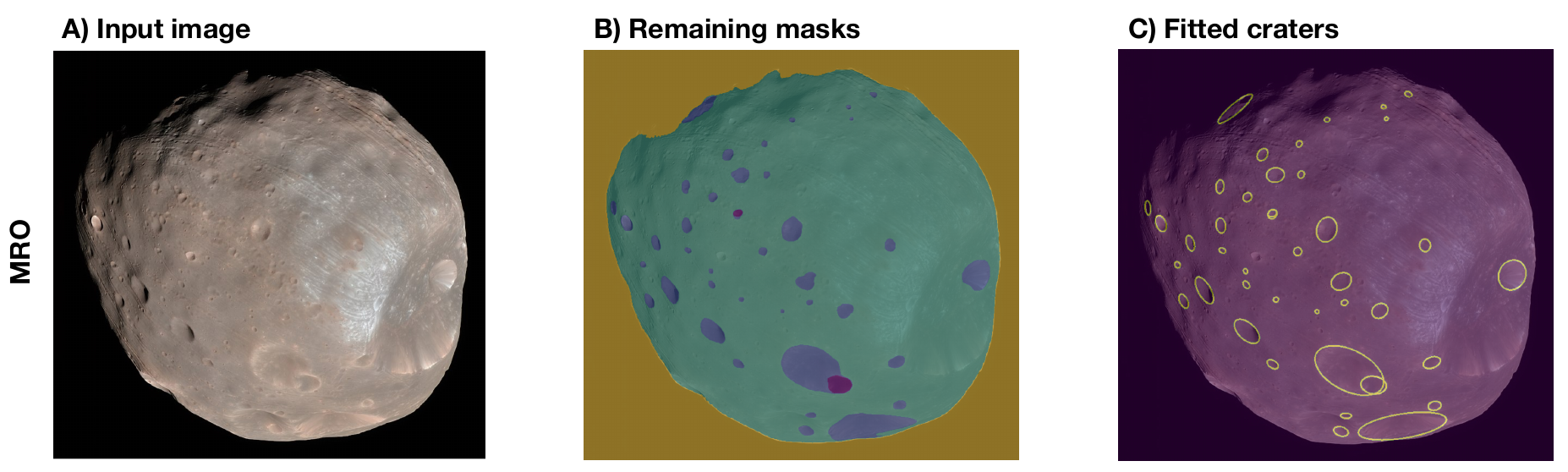}
\centering
\caption{The proposed CDA applied to Phobos using a false color image from Mars Reconnaissance Orbiter (MRO). A) column is the input image, B) are the remaining masks after filtering non-circular/elliptical shapes, and C) are the final fitted craters. Despite the unique nature of the input image, the proposed CDA works reasonably well, indicating the universality of SAM-based crater detection and pattern recognition in general.}
\label{F6}
\end{figure}

\subsubsection{Circular-Elliptical Indexes}
In the previous section SAM was applied to extract the segmentation masks of different morphological features for an input image. Since the majority of impact craters have a circular-elliptical shape \cite{Melosh:1989}, it is a rational choice to filter out all the segmentation masks that are not circular-elliptical. To do that we need to define a circularity and elliptical index, based on which the circularity-ellipticity of each mask will be assessed. 

Regarding circularity, if the mask is a circle then its radius ($r$) can be inferred from the measured area ($A$) (number of pixels) via $r=\sqrt{\frac{A}{\pi}}$. Consequently, its circumference can be calculated using the estimated radius $d=2\pi\sqrt{\frac{A}{\pi}}$. Subsequently, the perimeter of the mask ($P$) is calculated manually from the image, and if the shape is a circle then the ratio $n=d/P$ should be close to 1. This is a mainstream approach for calculating the circularity of an object \cite{Bottema:2000} with minimum computational requirements. One drawback of this approach is that elliptical shapes with low eccentricity can result in $n\approx 1$ and therefore give the false impression that an ellipse is a circle. To overcome this, we fit an ellipse to the investigated mask, and we define the index $m=\frac{a}{b}$ where $a$ and $b$ are the major axes of the ellipse. If both $n$ and $m$ equal $1\pm T$, then the shape is classified as a circle. The threshold $T$ is to be tuned depending on the type of the image, but from our experience a value $T\approx 0.1-0.5$ should be considered as a default. 

For the ellipticity, first we fit an ellipse to the investigated mask, and subsequently we infer its area from $w=\pi a b$, where $a$ and $b$ are the main axis of the fitted ellipse. Then we estimate the area of the mask ($A$) manually by measuring the pixels of the mask. If the mask is an ellipse then the ratio $q=\frac{w}{A}$ should be $q\approx 1\pm T$. The threshold $T$ is to be tuned but a value $T\approx 0.1-0.5$ should be considered as a default. Since by definition a circle is also an ellipse with $a=b$, in order to distinguish between circular and elliptical objects we first evaluate if an image is a circle (using $m$ and $n$) and if not only then do we further check the ellipticity index $q$. Lastly, additional constraints in the eccentricity and the ratio of $a/b$ can be trivially implemented to filter out elongated elliptical features.

Using $m,n$ and $q$ we filter out all the segmentation masks that are not circular-elliptical. The remaining masks are classified as circles or ellipses and a Canny filter is applied separately to each one of them to derive their edges. Lastly, a circle or an ellipse (depending on the mask classification) is fitted to the edges. The axes $a,b$, and the coordinates of the center are saved for each of the remaining masks. 

Figure \ref{F2} illustrates the results of applying the proposed CDA in Mars Express HRSC natural colour image of \textit{Ocrus Patera} (see Fig. \ref{F1}). It is indicative that the majority of the craters are correctly identified and mapped with a small amount of false positives. Even non-conventional (speculated \cite{Kolk:2001}) craters like the well-known elongated elliptical depression in the middle are correctly identified and sufficiently mapped. The undetected small craters in Fig. \ref{F2} are due to resolution constraints, and as it is shown in Fig. \ref{F3}, if we zoom in an investigated area the majority of the small craters will be correctly identified and mapped.

\section{Case Studies}

The SAM-based CDA is not constrained to a specific type of data, measurement setup and celestial body. This is because the core element of the proposed CDA is SAM, a generic foundation model for segmentation \cite{Kirillov:2023}, not fine-tuned for specific type of data. To demonstrate the universality of the proposed scheme we examine a set of case studies from different celestial bodies using different types of data and measurement setups. 

Figure \ref{F4} shows three case studies implementing the proposed CDA with Lunar data. The dataset comprised a DEM and two Lunar Reconnaissance Orbiter Camera (LROC) images captured from different angles. The outcomes of the experiment revealed that the proposed methodology could successfully identify and provide reasonably accurate estimations of the size of craters. Nonetheless, as noted in the previous section, the inability to identify smaller craters can be attributed to the resolution limitations inherent in SAM. However, this limitation can be mitigated by zooming into the image, as depicted in Figure \ref{F3}. Interestingly, as visible in the bottom panel of Figure 4, the algorithm is able to detect craters even in a high oblique angle image captured by LROC, thus suggesting the possibility of real-time detection of small craters using operational rover and lander cameras.

The second case study involves data from Mars. Figure \ref{F5} showcases three examples using Thermal Emission Imaging System (THEMIS) infrared images, High Resolution Imaging Experiment (HiRISE) orthoimages and a mosaic from European Space Agency's (ESA) HRSC. Similar to the previous case studies, the proposed CDA manages to both detect and measure the investigated craters with sufficient accuracy. SAM-based CDA work equally well regardless of the investigated celestial body indicating its potential for use in future space exploration missions.

In the last example we examine Phobos, the largest of the two Martian moons. We use a false colour image taken from Mars Reconnaissance Orbiter (MRO). Figure \ref{F6} shows the input, the remaining masks after filtering non-circular/elliptical shapes, and finally the fitted ellipses. The results are sufficiently good despite the fact that no specific training was done for this unique type of data. The proposed CDA manage to identify most of the craters with minimum false positives and negatives, despite the fact that it has not been trained for false colour MRO images.

\section{Discussion: Limitations and Future Work}

\begin{figure}[t]
\includegraphics[width=15cm]{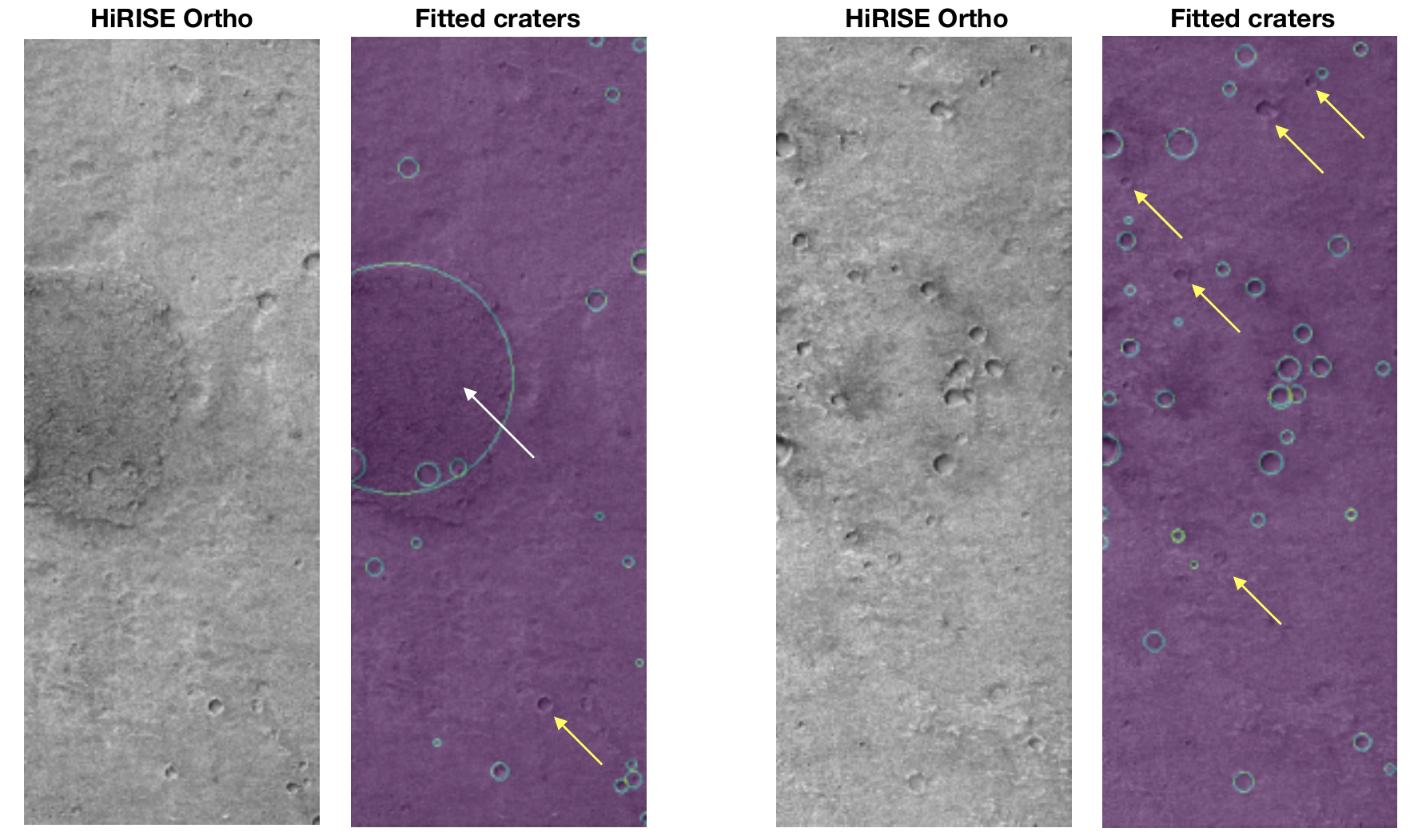}
\centering
\caption{Two examples using HiRISE orthoimages from Mars. With white arrow is a false positive due to the circular elevated topography that is wrongly classified as crater. Yellow arrows highlight false negatives on craters similar to the ones that the algorithm manage to correctly detect in the same image.}
\label{F7}
\end{figure}

The proposed SAM-based CDA is essentially a shape detector that focuses on circular/elliptical shapes. This is both an advantage and a drawback. This generic detection objective allows for the algorithm to work equally well despite the dataset and the investigated celestial body. SAM is trained using millions of images to identify segments and masks, and any mask that has a circular/elliptical shape will be identified as a crater. At the same time, this can give rise to false positives, since not all circular/elliptical shapes are craters. One common artefact that we encountered was that the central peaks of some craters were falsely identified as craters due to their circular shape. This can be easily overcome by adding an additional filter that removes craters with similar centers. However, this can also potentially remove any actual crater with its center coinciding the the larger crater leading to false negatives. Another typical artefact was miss-classification of shadows as craters due to their elliptical shape. These artefacts were filtered out using eccentricity thresholds not allowing elongated ellipses to be categorised as craters. Nonetheless, via this threshold the algorithm will not be able to detect actual elongated and elliptical craters. Another example of non-crater circular morphological feature is shown with white arrow in Figure \ref{F7}. The circular elevated topography seen in the image is falsely identified as crater due to its shape and distinct features. In the same Figure there are also some false negatives that are highlighted with yellow arrows. The algorithm failed to detect them despite detecting similar craters in the near proximity of the false negatives. This is an unexpected behaviour, which indicates that more research is needed to properly assess the limitations and instabilities of the proposed scheme. Another drawback of the proposed scheme is the need for tuning the threshold for the geometrical indexes. Depending on the data the circularity/elliptical threshold should be tuned accordingly in order to filter out non-crater masks. In noisy, low resolution and cluttered data, the threshold should be relaxed, which will lead to false positives. 

The above highlight that CDAs based on generic segmentation models (such as SAM) require further research and improvements. SAM is a generic foundation model that can be fine-tuned for planetary surfaces via transfer learning. Transfer learning refers to a variety of techniques aimed at using an existing pre-trained model on one task to perform another related task \cite{Sinno:2010}. Transfer learning has been successfully applied in geophysics where state of the art foundation models such as YOLO v3 \cite{yolov3}, pre-trained in big datasets (such as Imagenet \cite{Imagenet}), were used to further learn to detect specific geophysical targets of interest  \cite{Jesper:2018, Xiaofeng:2022}. Similar approaches, where a foundation pre-trained model is further trained for a specific task, have been widely applied in various scientific fields utilising the core capabilities of a foundation model combined with domain knowledge from a specialised well-labeled dataset \cite{Wang:2022,Xia:2021,Chiba:2021,Minoofam:2021}. For future work, the proposed SAM-based CDA could be potentially improved by using SAM as the foundation model and further train it with a diverse well-labelled dataset from various celestial bodies and different types of planetary data.

\section{Conclusions}
Through a series of examples using  different types of data from various celestial bodies, we demonstrated the potential of SAM as a CDA and a pattern recognition planetary tool in general. The proposed CDA performs equally well regardless of the type of the dataset and the celestial body, and without the need for additional labelled data for fine-tuning SAM. The current work lays the foundations for a single universal CDA for planetary science; and also introduces SAM as an effective way to identify patterns and dominant features in planetary data.  

\bibliographystyle{unsrt}  
\bibliography{ms}

\begin{thebibliography}{10}

\bibitem{Melosh:1989}
H.~J. {Melosh}.
\newblock {\em Impact Cratering}.
\newblock Oxford University Press, 1 edition, 1989.

\bibitem{Salamunicar:2012}
Goran Salamunićcar, Sven Lončarić, and Erwan Mazarico.
\newblock Lu60645gt and ma132843gt catalogues of lunar and martian impact
  craters developed using a crater shape-based interpolation crater detection
  algorithm for topography data.
\newblock {\em Planetary and Space Science}, 60(1):236--247, 2012.
\newblock Titan Through Time: A Workshop on Titan’s Formation, Evolution and
  Fate.

\bibitem{McSween:2019}
H.Y. McSween, J.E. Moersch, D.M. Burr, W.M. Dunne, J.P. Emery, L.C. Kah, and
  M.C. McCanta.
\newblock {\em Planetary Geoscience}.
\newblock Cambridge University Press, 2019.

\bibitem{Lemelin:2019}
Myriam Lemelin, Paul~G. Lucey, Katarina Miljković, Lisa~R. Gaddis, Trent Hare,
  and Makiko Ohtake.
\newblock The compositions of the lunar crust and upper mantle: Spectral
  analysis of the inner rings of lunar impact basins.
\newblock {\em Planetary and Space Science}, 165:230--243, 2019.

\bibitem{Huang:2018}
Jun Huang, Zhiyong Xiao, Jessica Flahaut, Mélissa Martinot, James Head, Xiao
  Xiao, Minggang Xie, and Long Xiao.
\newblock Geological characteristics of von kármán crater, northwestern south
  pole-aitken basin: Chang'e-4 landing site region.
\newblock {\em Journal of Geophysical Research: Planets}, 123(7):1684--1700,
  2018.

\bibitem{Hartman:2001}
G.~{Neukum} W.~K.~{Hartman}.
\newblock Cratering chronology and the evolution of mars.
\newblock {\em Space Science Reviews}, 96:165--194, 2001.

\bibitem{Yang:2020}
C.~Yang, H.~Zhao, H.~Bruzzone, J.~A. Benediktsson, Y.~Liang, B.~{Liu}, X.~Zeng,
  R.~Guan, C.~Li, and Z.~Ouyang.
\newblock Lunar impact crater identification and age estimation with chang'e
  data by deep and transfer learning.
\newblock {\em Nature Communications}, 11:6358, 2020.

\bibitem{Glaser:2014}
P.~Gläser, F.~Scholten, D.~{De Rosa}, R.~{Marco Figuera}, J.~Oberst,
  E.~Mazarico, G.A. Neumann, and M.S. Robinson.
\newblock Illumination conditions at the lunar south pole using high resolution
  digital terrain models from lola.
\newblock {\em Icarus}, 243:78--90, 2014.

\bibitem{Emami:2019}
Ebrahim Emami, Touqeer Ahmad, George Bebis, Ara Nefian, and Terry Fong.
\newblock Crater detection using unsupervised algorithms and convolutional
  neural networks.
\newblock {\em IEEE Transactions on Geoscience and Remote Sensing},
  57(8):5373--5383, 2019.

\bibitem{Downes:2020}
Lena Downes, Ted~J. Steiner, and Jonathan~P. How.
\newblock {\em Deep Learning Crater Detection for Lunar Terrain Relative
  Navigation}.
\newblock 2020.

\bibitem{Silvestrini:2022}
Stefano Silvestrini, Margherita Piccinin, Giovanni Zanotti, Andrea Brandonisio,
  Ilaria Bloise, Lorenzo Feruglio, Paolo Lunghi, Michèle Lavagna, and Mattia
  Varile.
\newblock Optical navigation for lunar landing based on convolutional neural
  network crater detector.
\newblock {\em Aerospace Science and Technology}, 123:107503, 2022.

\bibitem{Grant:2018}
John~A. Grant, Matthew~P. Golombek, Sharon~A. Wilson, Kenneth~A. Farley, Ken~H.
  Williford, and Al~Chen.
\newblock The science process for selecting the landing site for the 2020 mars
  rover.
\newblock {\em Planetary and Space Science}, 164:106--126, 2018.

\bibitem{Silburt:2018}
Ari Silburt, Mohamad Ali-Dib, Chenchong Zhu, Alan Jackson, Diana Valencia,
  Yevgeni Kissin, Daniel Tamayo, and Kristen Menou.
\newblock Lunar crater identification via deep learning.
\newblock {\em https://arxiv.org/abs/1803.02192v3}, 2018.

\bibitem{Robbins:2014}
Stuart~J. Robbins, Irene Antonenko, Michelle~R. Kirchoff, Clark~R. Chapman,
  Caleb~I. Fassett, Robert~R. Herrick, Kelsi Singer, Michael Zanetti, Cory
  Lehan, Di~Huang, and Pamela~L. Gay.
\newblock The variability of crater identification among expert and community
  crater analysts.
\newblock {\em Icarus}, 234:109--131, 2014.

\bibitem{Emami:2015}
Ebrahim Emami, George Bebis, Ara Nefian, and Terry Fong.
\newblock Automatic crater detection using convex grouping and convolutional
  neural networks.
\newblock In George Bebis, Richard Boyle, Bahram Parvin, Darko Koracin, Ioannis
  Pavlidis, Rogerio Feris, Tim McGraw, Mark Elendt, Regis Kopper, Eric Ragan,
  Zhao Ye, and Gunther Weber, editors, {\em Advances in Visual Computing},
  pages 213--224, Cham, 2015. Springer International Publishing.

\bibitem{Wetzler:2005}
P.~G. Wetzler, R.~Honda, B.~Enke, W.~J. Merline, C.~R. Chapman, and M.~C. Burl.
\newblock Learning to detect small impact craters.
\newblock In {\em 2005 Seventh IEEE Workshops on Applications of Computer
  Vision (WACV/MOTION'05) - Volume 1}, volume~1, pages 178--184, 2005.

\bibitem{Salamuniccar:2011}
Goran Salamunićcar, Sven Lončarić, Pedro Pina, Lourenço Bandeira, and José
  Saraiva.
\newblock Ma130301gt catalogue of martian impact craters and advanced
  evaluation of crater detection algorithms using diverse topography and image
  datasets.
\newblock {\em Planetary and Space Science}, 59(1):111--131, 2011.

\bibitem{Di:2014}
Kaichang Di, Wei Li, Zongyu Yue, Yiwei Sun, and Yiliang Liu.
\newblock A machine learning approach to crater detection from topographic
  data.
\newblock {\em Advances in Space Research}, 54(11):2419--2429, 2014.

\bibitem{Klear:2018}
P.~G. Wetzler, R.~Honda, B.~Enke, W.~J. Merline, C.~R. Chapman, and M.~C. Burl.
\newblock 9th planetary crater consortium.
\newblock In {\em An open-source library for automated crater detection},
  volume~1, 2018.

\bibitem{Lee:2019}
Christopher Lee.
\newblock Automated crater detection on mars using deep learning.
\newblock {\em Planetary and Space Science}, 170:16--28, 2019.

\bibitem{Kirillov:2023}
Alexander Kirillov, Eric Mintun, Nikhila Ravi, Hanzi Mao, Chloe Rolland, Laura
  Gustafson, Tete Xiao, Spencer Whitehead, Alexander~C. Berg, Wan-Yen Lo, Piotr
  Dollár, and Ross Girshick.
\newblock Segment anything.
\newblock {\em https://arxiv.org/abs/2304.02643v1}, 2023.

\bibitem{Szeliski:2011}
Richard Szeliski.
\newblock Computer vision algorithms and applications.
\newblock In {\em Computer vision algorithms and applications}. Springer, 2011.

\bibitem{Dhanachandra:2015}
Nameirakpam Dhanachandra, Khumanthem Manglem, and Yambem~Jina Chanu.
\newblock Image segmentation using k -means clustering algorithm and
  subtractive clustering algorithm.
\newblock {\em Procedia Computer Science}, 54:764--771, 2015.
\newblock Eleventh International Conference on Communication Networks, ICCN
  2015, August 21-23, 2015, Bangalore, India Eleventh International Conference
  on Data Mining and Warehousing, ICDMW 2015, August 21-23, 2015, Bangalore,
  India Eleventh International Conference on Image and Signal Processing, ICISP
  2015, August 21-23, 2015, Bangalore, India.

\bibitem{Qin:2011}
Kun Qin, Kai Xu, Feilong Liu, and Deyi Li.
\newblock Image segmentation based on histogram analysis utilizing the cloud
  model.
\newblock {\em Computers and Mathematics with Applications}, 62(7):2824--2833,
  2011.
\newblock Computers and Mathematics in Natural Computation and Knowledge
  Discovery.

\bibitem{Yi:2007}
Yi~Ma, Harm Derksen, Wei Hong, and John Wright.
\newblock Segmentation of multivariate mixed data via lossy data coding and
  compression.
\newblock {\em IEEE Transactions on Pattern Analysis and Machine Intelligence},
  29(9):1546--1562, 2007.

\bibitem{Farabet:2013}
Clement Farabet, Camille Couprie, Laurent Najman, and Yann LeCun.
\newblock Learning hierarchical features for scene labeling.
\newblock {\em IEEE Transactions on Pattern Analysis and Machine Intelligence},
  35(8):1915--1929, 2013.

\bibitem{Chen:2018}
Liang-Chieh Chen, George Papandreou, Iasonas Kokkinos, Kevin Murphy, and
  Alan~L. Yuille.
\newblock Deeplab: Semantic image segmentation with deep convolutional nets,
  atrous convolution, and fully connected crfs.
\newblock {\em IEEE Transactions on Pattern Analysis and Machine Intelligence},
  40(4):834--848, 2018.

\bibitem{Kim:2021}
Hyunwoo Kim, Jeonghoon Kim, Jungwook Choi, Jungkeol Lee, and Yong~Ho Song.
\newblock Binarized encoder-decoder network and binarized deconvolution engine
  for semantic segmentation.
\newblock {\em IEEE Access}, 9:8006--8027, 2021.

\bibitem{Kexin:2020}
Liu Kexin and Guo Chenjun.
\newblock Application of generative adversarial network in semantic
  segmentation.
\newblock In {\em 2020 17th International Computer Conference on Wavelet Active
  Media Technology and Information Processing (ICCWAMTIP)}, pages 343--348,
  2020.

\bibitem{Yang:2018}
Wenbin Yang, Quan Zhou, Jingnan Lu, Xiaofu Wu, Suofei Zhang, and Longin~Jan
  Latecki.
\newblock Dense deconvolutional network for semantic segmentation.
\newblock In {\em 2018 25th IEEE International Conference on Image Processing
  (ICIP)}, pages 1573--1577, 2018.

\bibitem{Noh:2015}
Hyeonwoo Noh, Seunghoon Hong, and Bohyung Han.
\newblock Learning deconvolution network for semantic segmentation.
\newblock In {\em 2015 IEEE International Conference on Computer Vision
  (ICCV)}, pages 1520--1528, 2015.

\bibitem{Buscombe:2022}
D.~Buscombe and E.~B. Goldstein.
\newblock A reproducible and reusable pipeline for segmentation of
  geoscientific imagery.
\newblock {\em Earth and Space Science}, 9(9):e2022EA002332, 2022.

\bibitem{Chen:2020}
Zhuoheng Chen, Xiaojun Liu, Jijin Yang, Edward Little, and Yu~Zhou.
\newblock Deep learning-based method for sem image segmentation in mineral
  characterization, an example from duvernay shale samples in western canada
  sedimentary basin.
\newblock {\em Computers and Geosciences}, 138:104450, 2020.

\bibitem{Collins:2020}
Adam~M. Collins, Katherine~L. Brodie, Andrew~Spicer Bak, Tyler~J. Hesser,
  Matthew~W. Farthing, Jonghyun Lee, and Joseph~W. Long.
\newblock Bathymetric inversion and uncertainty estimation from synthetic
  surf-zone imagery with machine learning.
\newblock {\em Remote Sensing}, 12(20), 2020.

\bibitem{Gupta:2021}
Ananya Gupta, Simon Watson, and Hujun Yin.
\newblock Deep learning-based aerial image segmentation with open data for
  disaster impact assessment.
\newblock {\em Neurocomputing}, 439:22--33, 2021.

\bibitem{Zhang:2018}
Zhengxin Zhang, Qingjie Liu, and Yunhong Wang.
\newblock Road extraction by deep residual u-net.
\newblock {\em IEEE Geoscience and Remote Sensing Letters}, 15(5):749--753,
  2018.

\bibitem{Liu:2023}
M.~Liu, J.~Liu, and X.~Ma.
\newblock Mrisnet: Deep-learning-based martian instance segmentation against
  blur.
\newblock {\em Earth Science Informatics}, 16:965--981, 2023.

\bibitem{Goh:2022}
Edwin Goh, Jingdao Chen, and Brian Wilson.
\newblock Mars terrain segmentation with less labels.
\newblock In {\em 2022 IEEE Aerospace Conference (AERO)}, pages 1--10, 2022.

\bibitem{Liu:2023_b}
Haiqiang Liu, Meibao Yao, Xueming Xiao, and Hutao Cui.
\newblock A hybrid attention semantic segmentation network for unstructured
  terrain on mars.
\newblock {\em Acta Astronautica}, 204:492--499, 2023.

\bibitem{Sofiiuk:2022}
Konstantin Sofiiuk, Ilya~A. Petrov, and Anton Konushin.
\newblock Reviving iterative training with mask guidance for interactive
  segmentation.
\newblock In {\em 2022 IEEE International Conference on Image Processing
  (ICIP)}, pages 3141--3145, 2022.

\bibitem{Qin:2022}
Qin Liu, Zhenlin Xu, Gedas Bertasius, and Marc Niethammer.
\newblock Simpleclick: Interactive image segmentation with simple vision
  transformers.
\newblock {\em Computers and Geosciences}, arXiv:2210.11006, 2022.

\bibitem{Lin:2014}
Tsung-Yi Lin, Michael Maire, Serge Belongie, James Hays, Pietro Perona, Deva
  Ramanan, Piotr Doll{\'a}r, and C~Lawrence Zitnick.
\newblock Microsoft coco: Common objects in context.
\newblock In {\em European Conference on Computer Vision}. Springer, 2014.

\bibitem{Gupta:2019}
Agrim Gupta, Piotr Dollar, and Ross Girshick.
\newblock Lvis: A dataset for large vocabulary instance segmentation.
\newblock In {\em Proceedings of the IEEE Conference on Computer Vision and
  Pattern Recognition}, 2019.

\bibitem{He:2022}
Kaiming He, Xinlei Chen, Saining Xie, Yanghao Li, Piotr Dollár, and Ross
  Girshick.
\newblock Masked autoencoders are scalable vision learners.
\newblock In {\em 2022 IEEE/CVF Conference on Computer Vision and Pattern
  Recognition (CVPR)}, pages 15979--15988, 2022.

\bibitem{Dosovitskiy:2021}
Alexey Dosovitskiy, Lucas Beyer, Alexander Kolesnikov, Dirk Weissenborn,
  Xiaohua Zhai, Thomas Unterthiner, Mostafa Dehghani, Matthias Minderer, Georg
  Heigold, Sylvain Gelly, Jakob Uszkoreit, and Neil Houlsby.
\newblock An image is worth 16x16 words: Transformers for image recognition at
  scale.
\newblock In {\em International Conference on Learning Representations}, 2021.

\bibitem{Bottema:2000}
M.J. Bottema.
\newblock Circularity of objects in images.
\newblock In {\em 2000 IEEE International Conference on Acoustics, Speech, and
  Signal Processing. Proceedings (Cat. No.00CH37100)}, volume~4, pages
  2247--2250 vol.4, 2000.

\bibitem{Kolk:2001}
D.A. van~der Kolk, K.L. Tribbett, E.B. Grosfils, S.E.H. Sakimoto, C.V.
  Mendelson, and J.E. Bleacher.
\newblock Orcus patera, mars: Impact crater or volcanic caldera?
\newblock In {\em Lunar and Planetary Science Conference}, 2001.

\bibitem{Sinno:2010}
Sinno~Jialin Pan and Qiang Yang.
\newblock A survey on transfer learning.
\newblock {\em IEEE Transactions on Knowledge and Data Engineering},
  22(10):1345--1359, 2010.

\bibitem{yolov3}
Joseph Redmon and Ali Farhadi.
\newblock Yolov3: An incremental improvement.
\newblock {\em arXiv}, 2018.

\bibitem{Imagenet}
Jia Deng, Wei Dong, Richard Socher, Li-Jia Li, Kai Li, and Li~Fei-Fei.
\newblock Imagenet: A large-scale hierarchical image database.
\newblock In {\em 2009 IEEE conference on computer vision and pattern
  recognition}, pages 248--255. Ieee, 2009.

\bibitem{Jesper:2018}
Jesper~S. Dramsch and Mikael Lüthje.
\newblock {\em Deep-learning seismic facies on state-of-the-art CNN
  architectures}, pages 2036--2040.
\newblock SEG, 2018.

\bibitem{Xiaofeng:2022}
Xiaofeng Li, Hai Liu, Feng Zhou, Zhongchang Chen, Iraklis Giannakis, and Evert
  Slob.
\newblock Deep learning–based nondestructive evaluation of reinforcement bars
  using ground-penetrating radar and electromagnetic induction data.
\newblock {\em Computer-Aided Civil and Infrastructure Engineering},
  37(14):1834--1853, 2022.

\bibitem{Wang:2022}
Xin Wang, Shuang Liu, and Changcai Zhou.
\newblock Classification of knee osteoarthritis based on transfer learning
  model and magnetic resonance images.
\newblock In {\em 2022 International Conference on Machine Learning, Control,
  and Robotics (MLCR)}, pages 67--71, 2022.

\bibitem{Xia:2021}
Xia Sun, Chengcheng Fu, Suoqi Liu, Wenjie Chen, Ran Zhong, Tingting He, and
  Xingpeng Jiang.
\newblock Multi-type microbial relation extraction by transfer learning.
\newblock In {\em 2021 IEEE International Conference on Bioinformatics and
  Biomedicine (BIBM)}, pages 266--269, 2021.

\bibitem{Chiba:2021}
Shohei Chiba and Hisayuki Sasaoka.
\newblock Basic study for transfer learning for autonomous driving in car race
  of model car.
\newblock In {\em 2021 6th International Conference on Business and Industrial
  Research (ICBIR)}, pages 138--141, 2021.

\bibitem{Minoofam:2021}
Seyyed Amir~Hadi Minoofam, Azam Bastanfard, and Mohammad~Reza Keyvanpour.
\newblock Trcla: A transfer learning approach to reduce negative transfer for
  cellular learning automata.
\newblock {\em IEEE Transactions on Neural Networks and Learning Systems},
  pages 1--10, 2021.

\end{thebibliography}

\end{document}